\documentclass[runningheads]{llncs}
\usepackage{graphicx}
\usepackage{amsmath}
\usepackage{amssymb}

\usepackage{changepage}
\usepackage[table, dvipsnames]{xcolor}
\definecolor{lightgray}{gray}{0.75}
\definecolor{lightergray}{gray}{0.95}

\newcommand{\etal}{\textit{et al}. }

\usepackage{cite}

\begin{document}
\title{Localized Region Contrast for Enhancing Self-Supervised Learning in Medical Image Segmentation}

\author{Xiangyi Yan\inst{1}, Junayed Naushad\inst{1,2}, Chenyu You\inst{3}, Hao Tang\inst{1}, Shanlin Sun\inst{1}, Kun Han\inst{1}, Haoyu Ma\inst{1}, James Duncan\inst{3}, Xiaohui Xie\inst{1}}
\authorrunning{Xiangyi et al.}

\institute{University of California, Irvine 
\\\email{\{xiangyy4, htang6, shanlins, kunh7, haoyum3, xhx\}@uci.edu}\\ 
\and University of Oxford 
\\\email{\{junayed.naushad\}@st-annes.ox.ac.uk}\
\and Yale University
\\\email{\{chenyu.you, james.duncan\}@yale.edu}\
}

\maketitle              
%
\begin{abstract}
Recent advancements in self-supervised learning have demonstrated that effective visual representations can be learned from unlabeled images. This has led to increased interest in applying self-supervised learning to the medical domain, where unlabeled images are abundant and labeled images are difficult to obtain. However, most self-supervised learning approaches are modeled as image level discriminative or generative proxy tasks, which may not capture the finer level representations necessary for dense prediction tasks like multi-organ segmentation. In this paper, we propose a novel contrastive learning framework that integrates Localized Region Contrast (LRC) to enhance existing self-supervised pre-training methods for medical image segmentation. Our approach involves identifying Super-pixels by Felzenszwalb’s algorithm and performing local contrastive learning using a novel contrastive sampling loss. Through extensive experiments on three multi-organ segmentation datasets, we demonstrate that integrating LRC to an existing self-supervised method in a limited annotation setting significantly improves segmentation performance. Moreover, we show that LRC can also be applied to fully-supervised pre-training methods to further boost performance.

\keywords{Self-supervised Learning  \and Contrastive Learning \and Semantic Segmentation.}

\end{abstract}

\section{Introduction}

Multi-organ segmentation is a crucial step in medical image analysis that enables physicians to perform diagnosis, prognosis, and treatment planning. However, manual segmentation of large volume computed tomography (CT) and magnetic resonance (MR) images is time-consuming and prone to high inter-rater variability. In recent years, deep convolutional neural networks (CNNs) have achieved state-of-the-art performance on a wide range of segmentation tasks for natural images \cite{long2015fully, he2017mask}. However, in the medical domain, there is often a lack of labeled examples to optimally train a deep neural network from scratch \cite{you2023rethinking, you2022mine, you2022bootstrapping}. Since unlabeled medical images are comparatively easier to obtain in larger quantities, an alternative strategy is to perform self-supervised learning and generate pre-trained models from unlabeled datasets. Self-supervised learning involves automatically generating a supervisory signal from the data itself and learning a representation by solving a pretext task.

In computer vision, current self-supervised learning methods can be broadly divided into discriminative modeling and generative modeling. In earlier times, discriminative self-supervised pretext tasks are designed as rotation prediction \cite{komodakis2018unsupervised}, jigsaw solving \cite{noroozi2016unsupervised}, and relative patch location prediction \cite{doersch2015unsupervised}, etc. Recently, contrastive learning achieves great success, whose core idea is to attract different augmented views of the same image and repulse augmented views of different images. Based on this, MoCo \cite{mocov1} and SimCLR \cite{chen2020simple} are proposed, which greatly shrink the gap between self-supervised learning and fully-supervised learning. More advanced techniques have emerged recently \cite{mocov2, mocov3, grill2020bootstrap}. 

Contrastive learning frameworks have also shown promising results in the medical domain, achieving good performance with few labeled examples \cite{chaitanya2020contrastive, azizi2021}.
Generative modeling also provides a feasible way for self-supervised pre-training \cite{zhang2017splitbrain, pathak2016context, zhang2016colorful}. Recently, He \etal propose MAE \cite{xiao2021early} and yield a nontrivial and meaningful generative self-supervisory task, by masking a high proportion of the input image. Transfer learning performance in downstream tasks outperforms supervised pre-training and shows promising scaling behavior. 

In medical image domain, Model Genesis\cite{models_genesis} uses a "painting" operation to generate a new image by modifying the input image. Several self-supervised learning approaches have also achieved state-of-the-art performance in the medical domain on both classification and segmentation tasks while significantly reducing annotation cost \cite{chaitanya2020contrastive, haghighi2021transferable}

However, most self-supervised pre-training strategies are image\cite{mocov1, swav, chen2020simple, gansbeke2021, bai2022} or patch\cite{chaitanya2020contrastive, azizi2021} level, which are not capable of capturing the detailed feature representations required for accurate medical segmentation. To address this issue, in this paper, we propose a novel contrastive learning framework that integrates Localized Region Contrast (LRC) to enhance existing self-supervised pre-training methods for medical image segmentation.

Our proposed framework leverages Felzenszwalb’s algorithm \cite{Felzenszwalb2004} to formulate local regions and defines a novel contrastive sampling loss to perform localized contrastive learning. Our main contributions include 
\begin{itemize}
    \item We propose a standalone localized contrastive learning module that can be integrated into most existing pre-training strategy to boost multi-organ segmentation performance by learning localized feature representations. 
    \item We introduce a novel localized contrastive sampling loss for dense self-supervised pre-training on local regions.
    \item We conduct extensive experiments on three multi-organ segmentation benchmarks and demonstrate that our method consistently outperforms current supervised and unsupervised pre-training approaches.
\end{itemize}

\section{Methodology}

\begin{figure}[h]

\includegraphics[width=\textwidth]{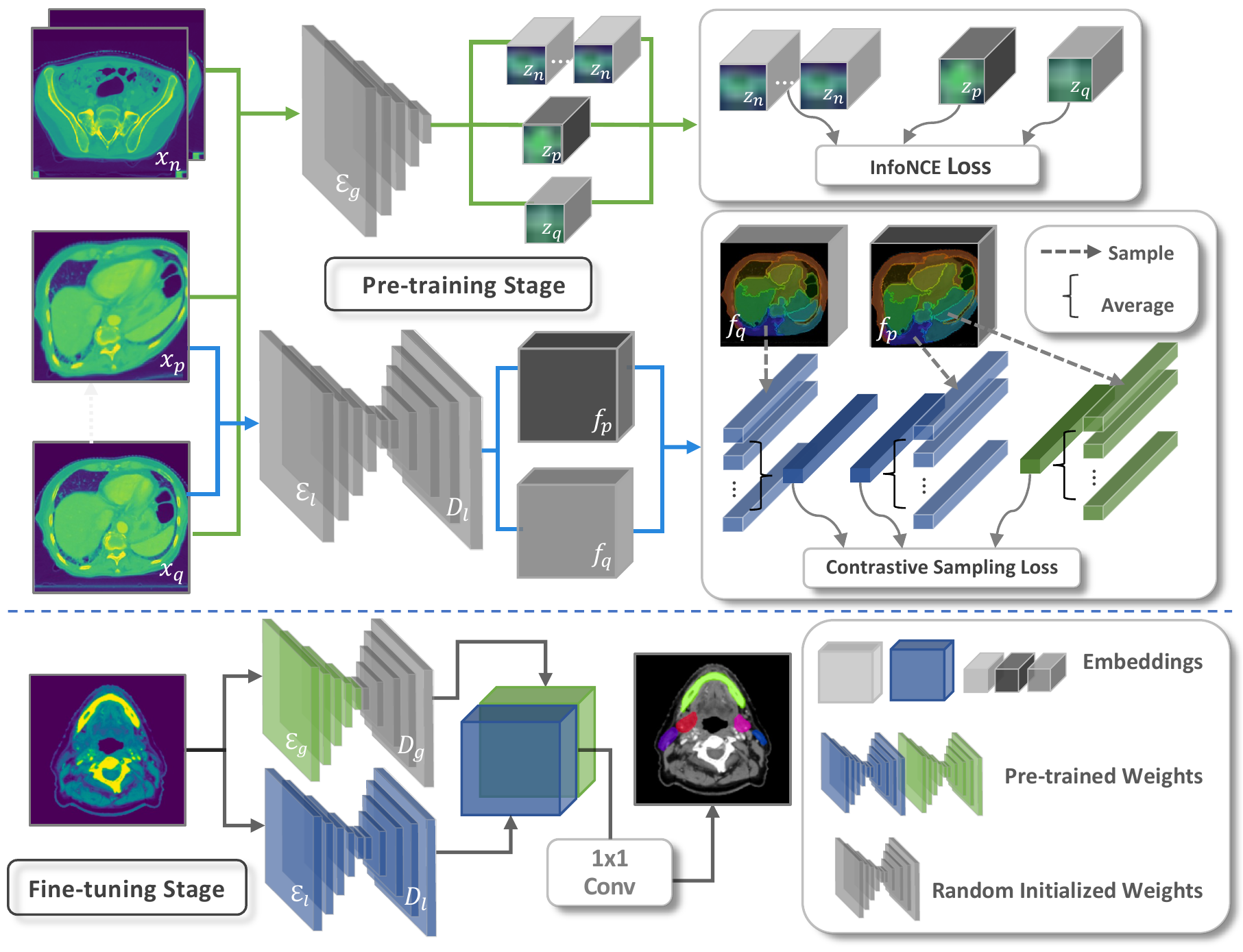}
\caption{Overview of our self-supervised framework with pre-training stage, consisting of global (with an example of MoCo) and our localized contrastive loss, and fine-tuning stage.} \label{fig1}
\end{figure}

Figure \ref{fig1} illustrates our complete framework, which comprises two stages: the contrastive pre-training stage and the fine-tuning stage. Although LRC can be integrated with most of the current popular pre-training strategies, for the purpose of illustration, in this section, we demonstrate how to integrate our LRC module with the classical global (image-level) contrast pre-training strategy MoCo\cite{mocov1}, using both its original global contrast and our localized contrastive losses during the contrastive pre-training stage. During the fine-tuning stage, we simply concatenate the local and global contrast models and fine-tune the resulting model on a small target dataset. Further details about each stage are discussed in the following subsections.

\subsection{Pre-training Stage}
In the pre-training stage, for each batch an image $\boldsymbol{x}_q$ is randomly chosen from $B$ images as a query sample, and the rest of the images $\boldsymbol{x}_n \in \{\boldsymbol{x}_1, \boldsymbol{x}_2, ..., \boldsymbol{x}_B\}$ are considered as negative key samples, where $n \neq q$. To formulate a positive key sample $\boldsymbol{x}_p$, elastic transforms are performed on the query sample $\boldsymbol{x}_q$. 

\textbf{Global Contrast} To explore global contextual information, we train a latent encoder $\mathcal{E}_g$ following the contrastive protocol in \cite{mocov1}. Three sets of latent embeddings $\boldsymbol{z}_q$, $\boldsymbol{z}_p$, $\boldsymbol{z}_n$ are extracted by $\mathcal{E}_g$ from $\boldsymbol{x}_q$, $\boldsymbol{x}_p$, $\boldsymbol{x}_n$ respectively. Using dot product as a measure of similarity, a form of a contrastive loss function called InfoNCE\cite{infonce} is considered:
$\mathcal{L}_{g}=-\log \frac{\exp \left(\boldsymbol{z}_q \cdot \boldsymbol{z}_p / \tau_g\right)}{\sum_{i=1}^{B} \exp \left(\boldsymbol{z}_q \cdot \boldsymbol{z}_i / \tau_g\right)}$, 
where $\tau_g$ is the global temperature hyper-parameter per \cite{tau}. Note that in the global contrast branch, we only pre-train $\mathcal{E}_g$.

\textbf{Local Region Contrast}
Unlike global contrast, positive and negative pairs for local contrast are only generated from input image $\boldsymbol{x}_q$ and its transform $\boldsymbol{x}_p$. We differentiate local regions and formulate the positive and negative pairs by using Felzenszwalb’s algorithm. For an input image $\boldsymbol{x}$, Felzenszwalb’s algorithm provides $K$ local regions  $\mathcal{R} = \{\boldsymbol{r}^1, \boldsymbol{r}^2, .., \boldsymbol{r}^{K}\}$, where $\boldsymbol{r}^{k}$ is the k-th local region cluster for image $\boldsymbol{x}$. We then perform elastic transform for both the query image $\boldsymbol{x}_q$ and its local regions $\mathcal{R}_q$ so that we have the augmented image $\boldsymbol{x}_p = T_e(\boldsymbol{x}_q)$ and its local regions $\mathcal{R}_p = \{\boldsymbol{r}_p^1, \boldsymbol{r}_p^2, .., \boldsymbol{r}_p^{K_p}\}$, where $\boldsymbol{r}_p^{k} = T_e(\boldsymbol{r}_q^k)$. Note that $K_q = K_p$ always holds since $\mathcal{R}_p$ is a one-to-one mapping from $\mathcal{R}_q$. Following the widely used U-Net \cite{ronneberger2015unet} model design, the query image $\boldsymbol{x}_q$ and augmented image $\boldsymbol{x}_p$ are then forwarded to a randomly initialized U-Net variant, which includes a convolutional encoder $\mathcal{E}_l$ and a convolutional decoder $\mathcal{D}_l$. 
We get corresponding feature maps $\boldsymbol{f}_q$ and $\boldsymbol{f}_p$ with the same spatial dimensions as $\boldsymbol{x}_q$ and $\boldsymbol{x}_p$ and $D$ channels from the last convolutional layer of $\mathcal{D}_l$. Afterwards, we sample $N$ vectors with dimension $D$ from the local region $\boldsymbol{r}_q^k$ in $\boldsymbol{f}_q$, and formulate the sample mean $\overline{{\boldsymbol{f}_{q}^{k}}} = \frac{1}{N}\sum_{n=1}^{N} \boldsymbol{f}_q^{k, n}$, where $\boldsymbol{f}_q^{k, n}$ is the n-th vector sampled from feature map $\boldsymbol{f}_{q}$ within the k-th local region $\boldsymbol{r}_q^k$. Our sampling strategy is straightforward: we sample random points with replacement following a uniform distribution. We simply refer to this as “random sampling”. Similarly, for feature map $\boldsymbol{f}_p$, its sample mean $\overline{{\boldsymbol{f}_{p}^{k}}}$ can be provided following the same random sampling process. Each local region pair of $\overline{{\boldsymbol{f}_{q}^{k}}}$ and $\overline{{\boldsymbol{f}_{p}^{k}}}$ is considered a positive pair. For the negative pairs, we sample both $\boldsymbol{f}_q$ and $\boldsymbol{f}_p$ from the rest of the local regions $\{\boldsymbol{r}^1, \boldsymbol{r}^2, ..., \boldsymbol{r}^{k-1}, \boldsymbol{r}^{k+1}, ..., \boldsymbol{r}^K\}$. The local contrastive loss can be defined as follows:
$$\mathcal{L}_{l}=-\sum_{k_1=1}^{K}\log \frac{\exp \left(\overline{{\boldsymbol{f}_{q}^{k_1}}} \cdot \overline{{\boldsymbol{f}_{p}^{k_1}}} / \tau_l\right)}{\sum_{k_2=1}^{K} \exp \left(\overline{{\boldsymbol{f}_{q}^{k_1}}} \cdot \overline{{\boldsymbol{f}_{q}^{k_2}}} / \tau_l\right) + \exp \left(\overline{{\boldsymbol{f}_{q}^{k_1}}} \cdot \overline{{\boldsymbol{f}_{p}^{k_2}}} / \tau_l\right)
}
$$
where $\tau_l$ is the local temperature hyper-parameter. Compared to the global contrast branch, in local contrastive learning, we pre-train both $\mathcal{E}_l$ and $\mathcal{D}_l$.


\subsection{Fine-tuning Stage}

In the former pre-training stages, $\mathcal{E}_g$, $\mathcal{E}_l$, and $\mathcal{D}_l$ are pre-trained with global and local contrast strategy accordingly, with a large number of unlabelled images. In the fine-tuning stage, we fine-tune the model with a limited number of labelled images $x_f \in \{x_1, x_2, ..., x_F\}$, where $F$ is the size of the fine-tuning dataset. 
Besides the two pre-trained encoders and one decoder, a randomly initialized decoder $\mathcal{D}_g$ is appended to the pre-trained $\mathcal{E}_g$ to ensure that the embeddings have the same dimensions prior to concatenation.
We combine local and global contrast models by concatenating the output of $\mathcal{D}_g$ and $\mathcal{D}_l$'s last convolutional layer, and fine-tune on the target dataset in an end-to-end fashion. Different levels of feature maps from encoders are concatenated with corresponding layers of decoders through skip connections to provide alternative paths for the gradient. Dice loss is applied as in usual multi-organ segmentation tasks.


\section{Experimental Results}
\subsection{Pre-training Dataset} 
During both global and local pre-training stages, we pre-train the encoders on the Abdomen-1K \cite{Ma-2021-AbdomenCT-1K} dataset. It contains over one thousand CT images which equates to roughly 240,000 2D slices. The CT images have been curated from 12 medical centers and include multi-phase, multi-vendor, and multi-disease cases. Although segmentation masks for liver, kidney, spleen, and pancreas are provided in this dataset, we ignore these labels during pre-training since we are following the self-supervised protocol.

\subsection{Fine-tuning Datasets} 
During the fine-tuning stage, we perform extensive experiments on three datasets with respect to different regions of the human body.  

ABD-110 is an abdomen dataset from \cite{Tang_2021_ICCV} that contains 110 CT images from patients with various abdominal tumors and these CT images were taken during the treatment planning stage. We report the average DSC on 11 abdominal organs (large bowel, duodenum, spinal cord, liver, spleen, small bowel, pancreas, left kidney, right kidney, stomach and gallbladder).

Thorax-85 is a thorax dataset from \cite{thorax-dataset} that contains 85 thoracic CT images. We report the average DSC on 6 thoracic organs (esophagus, trachea, spinal cord, left lung, right lung, and heart).

HaN is from \cite{tang2019clinically} and contains 120 CT images covering the head and neck region. We report the average DSC on 9 organs (brainstem, mandible, optical chiasm, left optical nerve, right optical nerve, left parotid, right parotid, left submandibular gland, and right submandibular gland).

\subsection{Implementation Details}

All images are re-sampled to have spacing of 2.5mm $\times$ 1.0mm $\times$ 1.0mm, with respect to the depth, height, and width of the 3D volume. In the pre-training stage, we apply elastic transform to formulate positive samples. 
\begin{table}[!t]
\caption{Comparison of our proposed pre-training strategy combining local contrast with different pre-training methods. Models are fine-tuned on three datasets where $|X_{T}|$ is the number of labeled CT images, and the evaluation metric is Dice score. \textbf{Bold} numbers indicate corresponding global pre-training method is enhanced by LRC.}

\begin{center}
\begin{tabular}{|c|c|c|c|c|c|c|c|c|c|c|c|c|c|} \hline

\rowcolor{white} \multicolumn{2}{|c|}{Global Pre-} & \multicolumn{4}{c|}{ABD-110} & \multicolumn{4}{c|}{Thorax-85} & \multicolumn{4}{c|}{HaN} \\ \cline{3-14} 
\rowcolor{white} \multicolumn{2}{|c|}{training Method } & \multicolumn{2}{c|}{$|X_{T}|$=10} & \multicolumn{2}{c|}{$|X_{T}|$=60} & \multicolumn{2}{c|}{$|X_{T}|$=10} & \multicolumn{2}{c|}{$|X_{T}|$=60} & \multicolumn{2}{c|}{$|X_{T}|$=10} & \multicolumn{2}{c|}{$|X_{T}|$=60} \\ \cline{1-14}
\rowcolor{white} \multicolumn{2}{|c|}{w/ or wo/ LRC} & {w/o}& {w/} & {w/o} & {w/} & {w/o} & {w/} & {w/o} & {w/} & {w/o} & {w/} & {w/o} & {w/}  \\ 

\cline{1-14}

\rowcolor{white} \multicolumn{2}{|c|}{Random init} & 68.8 & \textbf{70.9} & 76.0 & \textbf{78.0} & 85.9 & \textbf{87.8} & 89 & \textbf{89.4} & 50.9 & \textbf{52.6} & 77.8 & \textbf{78.0}  \\ 

\cline{1-14}
\rowcolor{lightergray} \multicolumn{14}{|c|}{{{Supervised Pre-training on Natural Images}}} \\ 
\cline{1-14}

\rowcolor{white} \multicolumn{2}{|c|}{ImageNet\cite{russakovsky2015imagenet}} & 70.9 & \textbf{72.7} & 77.6 & \textbf{78.6} & 87.2 & \textbf{88.1} & 89.4 & \textbf{90.3} & 67.5 & \textbf{72.6} & 77.0 & \textbf{77.9}    \\ 

\cline{1-14}
\rowcolor{lightergray} \multicolumn{14}{|c|}{{{Discriminative Self-supervised Pre-training}}} \\ 
\cline{1-14}

\rowcolor{white} \multicolumn{2}{|c|}{Relative Loc\cite{doersch2015unsupervised}} & 69.1 & \textbf{72.6} & 76.4 & \textbf{78.0} & 86.1 & 86.0 & 89.4 & \textbf{89.6} & 55.2 & \textbf{60.3} & 77.9 & 77.9    \\ 

\rowcolor{white} \multicolumn{2}{|c|}{Rotation Pred\cite{komodakis2018unsupervised}} & 69.1 & \textbf{70.2} & 77.0 & \textbf{78.1} & 86.3 & 84.6 & 89.2 & \textbf{89.7} & 54.7 & \textbf{59.6} & 76.8 & \textbf{77.3}  \\

\rowcolor{white} \multicolumn{2}{|c|}{MoCo v1\cite{mocov1}} & 72.1 & \textbf{75.3} & 77.7 & \textbf{79.0} & 86.6 & \textbf{88.6} & 89.3 & \textbf{90.1} & 52.3 & \textbf{70.6} & 76.0 & \textbf{77.2}   \\ 

\rowcolor{white} \multicolumn{2}{|c|}{MoCo v2\cite{mocov2}} & 72.2 & \textbf{75.2} & 77.9 & \textbf{79.6} & 86.6 & \textbf{87.4} & 89.7 & \textbf{90.0} & 55.8 & \textbf{68.2} & 76.7 & \textbf{77.5}    \\ 

\rowcolor{white} \multicolumn{2}{|c|}{BYOL\cite{grill2020bootstrap}} & 71.6 & \textbf{74.8} & 78.0 & \textbf{79.0} & 87.3 & \textbf{88.2} & 89.2 & \textbf{89.5} & 53.1 & \textbf{61.1} & 76.3 & \textbf{76.6}   \\ 

\rowcolor{white} \multicolumn{2}{|c|}{DenseCL\cite{wang2021dense}} & 71.6 & \textbf{72.0} & 77.2 & \textbf{78.5} & 84.3 & \textbf{85.1} & 87.5 & \textbf{87.8} & 59.5 & \textbf{62.2} & 76.7 & 76.7   \\ 

\rowcolor{white} \multicolumn{2}{|c|}{SimSiam\cite{chen2021exploring}} & 73.4 & \textbf{76.0} & 79.2 & \textbf{79.5} & 88.2 & 87.0 & 88.6 & \textbf{89.9} & 57.2 & \textbf{63.2} & 78.9 & 77.2  \\ 

\cline{1-14}
\rowcolor{lightergray} \multicolumn{14}{|c|}{{{Generative Self-supervised Pre-training}}} \\ 
\cline{1-14}

\rowcolor{white} \multicolumn{2}{|c|}{Models Genesis\cite{models_genesis}} & 72.9 & \textbf{73.2} & 80.2 & \textbf{80.6} & 88.2 & \textbf{88.4} & 90.1 & \textbf{91.3} & 64.0 & \textbf{67.2} & 74.2 & 73.2   \\ 

\rowcolor{white} \multicolumn{2}{|c|}{MAE\cite{mae}} & 71.5 & 71.2 & 79.5 & \textbf{79.7} & 86.2 & \textbf{86.5} & 89.3 & 89.0 & 52.8 & \textbf{55.2} & 77.3 & \textbf{77.5}    \\

\hline 

\end{tabular}

\label{table:comparison_results}
\end{center}
\end{table}
In the global contrast branch, we use the SGD optimizer to pre-train a ResNet-50 \cite{he2016deep} (for MAE\cite{mae}, we use ViT-base\cite{dosovitskiy2020vit}.) encoder $\mathcal{E}_g$ for 200 epochs. In the local contrast branch, we use the Adam \cite{adam} optimizer to pre-train both encoder $\mathcal{E}_l$ and decoder $\mathcal{D}_l$ for 30 epochs. The dimension of sampled vectors $D$ is 64 since $\boldsymbol{f}_q$ and $\boldsymbol{f}_p$ have 64 channels. In the fine-tuning stage, we use the Adam optimizer to train the whole framework in an end-to-end fashion. All optimizers in both pre-training and fine-tuning stages are set to have momentum of 0.9 and weight decay of $10^{-4}$.

\subsection{Quantitative Results}
In table \ref{table:comparison_results}, we select 9 self-supervised pre-trained with 1 ImageNet supervised pre-trained networks and combine with our proposed localized region contrast (LRC). Through extensive experiments on 3 different datasets, we demonstrate LRC is capable of enhancing these pre-training algorithms in a consistent way. We use Sørensen–Dice coefficient (DSC) to measure our experimental results. 

For ABD-110, LRC enhances 9 and 10 out of 10 pre-training approaches, with $|X_{T}|=10$ and 60 respectively. 
For thorax-85, LRC enhances 7 and 9 out of 10 pre-training approaches, with $|X_{T}|=10$ and 60 respectively.
For HaN, LRC enhances 10 and 6 out of 10 pre-training approaches, with $|X_{T}|=10$ and 60 respectively.

The experiments consistently show LRC boosts the multi-organ segmentation performance of most global contrast model across all three datasets. 

\begin{figure}[h!]
\begin{center}
\includegraphics[width=290pt]{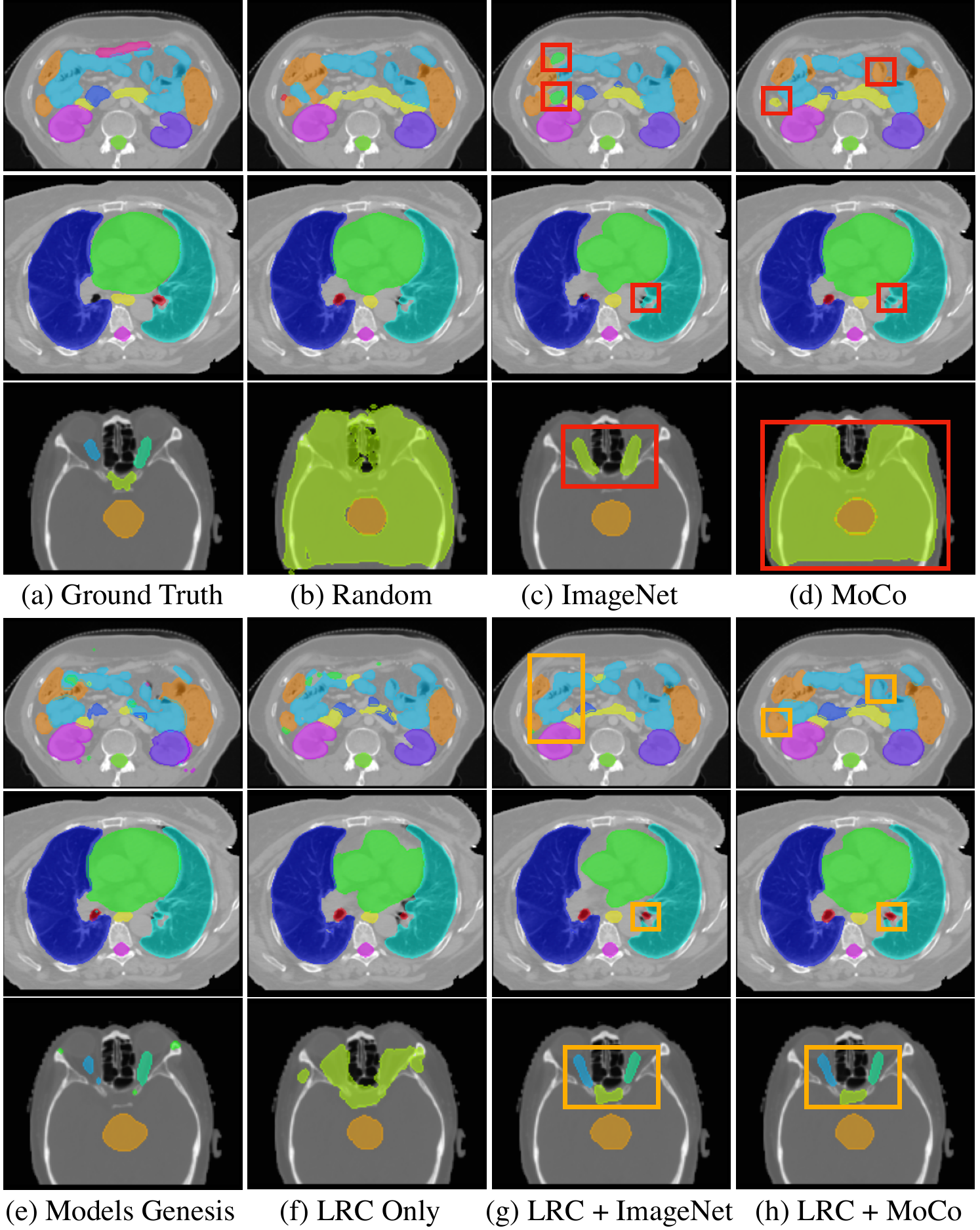}
\caption{Ground truth segmentation masks and predictions on a slice from each dataset. Due to limited space, we only demonstrate selected global pre-training methods. By comparing (c) with (g) and (d) with (h), our method shows significant improvement, particularly on the challenging HaN dataset.} \label{fig2}
\end{center}
\end{figure}

\subsection{Qualitative Results}
In figure \ref{fig2}, we show segmentation results on ABD-110, Thorax-85, and HaN datasets respectively. All the results are provided by models trained with target dataset size $|X_{T}|=10$. By comparing (c) with (g) and (d) with (h), our method shows significant improvement, particularly on the challenging HaN dataset. However, due to limited space, we are only capable of demonstrating selected global pre-training methods.

\subsection{Visualization of Localized Regions}
\begin{figure}[h!]

\includegraphics[width=\textwidth]{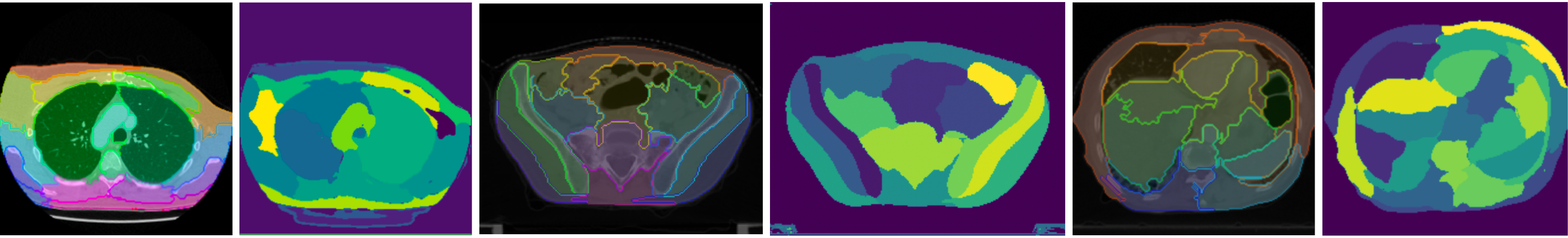}
\caption{Three pairs of examples comparing the local regions generated by Felzenszwalb's algorithm on the left and K-Means clustering of the embeddings from the local contrast model on the right.} \label{fig3}

\end{figure}

Figure \ref{fig3} presents three pairs of localized region visualizations generated by Felzenszwalb's algorithm (with a black background) and the corresponding feature representations extracted from LRC. We use K-Means clustering to formulate these feature representations into K clusters, which are shown in purple in the figure. Our results demonstrate that LRC learns informative semantic feature representations that can be effectively clustered using a simple K-Means algorithm.

\subsection{Ablation Study}
\subsubsection{Effect of additional parameters}

Additional parameters do bring performance enhancement in machine learning. However, in Table. \ref{table:add_para}, we show our proposed LRC enhances the performance of general global pre-training approaches by integrating localized features rather than simply adding additional parameters. We prove this argument by adding the same amount of MoCo pre-trained network parameters, to the above global pre-trained methods. As a result, LRC outperforms MoCo under every setting. In this experiment, we use ABD-110 as fine-tuning dataset and set $|X_{T}|=60$.

\begin{table}[!t]
\caption{Comparison of integrating LRC vs MoCo with different pre-training methods. We show LRC enhances global pre-training approaches by integrating localized features rather than simply adding additional parameters.}

\begin{center}
\begin{tabular}{|c|c|c|c|c|c|c|c|} \hline
\rowcolor{white} \multicolumn{2}{|c|}{Global Pre- } & {ImageNet} & {Relative} &{Rotation} & {MAE} & {BYOL} & {SimSiam} \\ 
\rowcolor{white} \multicolumn{2}{|c|}{training Method } & {\cite{russakovsky2015imagenet}} & {Loc\cite{doersch2015unsupervised}} & {Pred\cite{komodakis2018unsupervised}} & {\cite{mae}} & {\cite{grill2020bootstrap}} & {\cite{chen2021exploring}} \\ \cline{1-8}
\rowcolor{white} \multicolumn{2}{|c|}{w/ MoCo \cite{mocov1}} & 78.4 & 77.1 & 77.3 & 79.5 & 78.2 & \textbf{79.5}   \\ 
\rowcolor{white} \multicolumn{2}{|c|}{w/ LRC} & \textbf{78.6} & \textbf{78.0} & \textbf{78.1} & \textbf{79.7} & \textbf{79.0} & \textbf{79.5}\\ 
\hline
\end{tabular}
\label{table:add_para}
\end{center}
\end{table}

\subsubsection{Number of samples $N$}

In table \ref{table:num_sample}, we explore the effect of different number of samples $N$ to the contrastive sampling loss. When the sample mean $\overline{{\boldsymbol{f}^{k}}}$ is only averaged from a small number of vectors, the capability of representing certain region level can be limited. In the opposite, when the number of samples $N$ is large, the sampling bias can be high, since the number of pixels can be smaller than $N$. Therefore, we need a proper choice of $N$. With $N=50$, our method demonstrates the best DSC score of 0.732.

\begin{table}[h!]
\caption{Different number of samples $N$ largely influences the fine-tuning results. Results are provided by LRC + MoCo fine-tuned on ABD-110 with $|X_T|=$10.}

\centering
\begin{tabular}{c|c|c|c}
\hline
$N$         &       10          & 50 &  100  \\ \hline
DSC &     0.695 &     \textbf{0.732} &     0.717      \\ \hline
                       
\end{tabular}

\label{table:num_sample}
\end{table}


\section{Conclusion}


In this paper, we propose a contrastive learning framework, which integrates a novel localized contrastive sampling loss and enables the learning of fine-grained representations that are crucial for accurate segmentation of complex structures. Through extensive experiments on three multi-organ segmentation datasets, we demonstrated that our approach consistently boosts current supervised and unsupervised pre-training methods. LRC provides a promising direction for improving the accuracy of medical image segmentation, which is a crucial step in various clinical applications. Overall, we believe that our approach can significantly benefit the medical image analysis community and pave the way for future developments in self-supervised learning for medical applications.

\newpage
\bibliographystyle{splncs04}
\bibliography{bibfile}
\end{document}